\documentclass[a4paper]{article}

\usepackage{INTERSPEECH2019}
\usepackage[dvipsnames]{xcolor}
\usepackage{algorithm}
\usepackage{algorithmicx}
\usepackage{algpseudocode}
\usepackage{capt-of}
\usepackage{subfigure}

\algrenewcommand\algorithmicindent{1em}

\title{Efficient Dynamic WFST Decoding for Personalized Language Models}
\name{Jun Liu, Jiedan Zhu, Vishal Kathuria, Fuchun Peng}
\address{
	Facebook AI, Menlo Park, CA, USA}
\email{\{junliu, jiedan, vishal, fuchunpeng\}@fb.com}

\begin{document}
	
	\maketitle
	\begin{abstract}
		We propose a two-layer cache mechanism to speed up dynamic WFST decoding with personalized language models. The first layer is a public cache that stores most of the static part of the graph. This is shared globally among all users. A second layer is a private cache that caches the graph that represents the personalized language model, which is only shared by the utterances from a particular user. We also propose two simple yet effective pre-initialization methods, one based on breadth-first search, and another based on a data-driven exploration of decoder states using previous utterances. Experiments with a calling speech recognition task using a personalized contact list demonstrate that the proposed public cache reduces decoding time by factor of three compared to decoding without pre-initialization. 
		Using the private cache provides additional efficiency gains, reducing the decoding time by a factor of five.
		
	\end{abstract}
	
	
	\section{Introduction}
	
	Speech input is now a common feature for smart devices. In many cases, the user's query involves entities such as a name from a contact list, a location, or a music title. Recognizing entities is particularly challenging for speech recognition because many entities are infrequent or out of the main vocabulary of the system. One way to improve performance is such cases is through the use of a personal language model (LM) which contains the expected user-specific entities. Because each user can have their own personalized LM, it is vital that the speech decoder be able to efficiently load the model on the fly, so it can be used in decoding, without any noticeable increase in latency.  
	
	Many state-of-the-art speech recognition decoders are based on the weighted finite state transducer (WFST) paradigm~\cite{mohriwfst, mowfst}. A conventional WFST decoder searches a statically composed $H C L G$ graph, where $H$ is the graph that translates HMM states to CD phones, $C$ translates CD phones to graphemes, $L$ translates graphemes to words and $G$ is graph that represents the language model.  Using a statically composed graph has two limitations. First, it is both compute and memory intensive when the vocabulary and LM are large. Second, the static graph approach makes it hard to handle personalized language models~\cite{aleksic2015improved}. Many common tasks a user may want to perform with a voice assistant such as making phone calls, messaging to a specific contact or playing favorite music require a personalized language model.  A dynamic WFST decoder is better suited for such cases.  As denoted in Eq~\eqref{eqn:dynamic}, in a dynamic WFST decoder, $HCL$ is composed and optimized offline, while $G$ is composed on the fly with lazy (on-demand) composition, denoted as $\circ$.  
	\begin{equation}\label{eqn:dynamic}
	T = H  C  L \circ G
	\end{equation}
	
	To handle dynamic entities, a class LM $G_c$ is normally used as background $G$ and a personalized LM $G_p$ is replaced on-the-fly, before applying lazy composition. 
	\begin{equation}\label{eqn:replace}
	T = H C L \circ Replace(G_c, G_p)
	\end{equation}
	
	Since the non-terminal states are composed on-the-fly, it means the states of recognition FST will also contain personalized information that cannot be used by other users or service threads.
	
	In previous work, a method was proposed to do a pre-initialized composition for a non-class LM ~\cite{allauzen2013pre}. However, it the dynamic part is still expanded on-the-fly. In this work, we propose two improvements in order to best leverage class language models.  First, we use simpler methods for pre-initialization which do not need to pre-generate decoder state statistics. Second, we propose a two-layer pre-initialization mechanism that also avoids performing dynamic expansion on per user basis. In the two-layer pre-initialization method, we make use of a class LM with class tag. We build a personalized FST that contains the members of the class for each user. Using the FST replacement algorithm, we obtain a personalized language transducer~\cite{openfst}. We perform a pre-composition for all FST states whose transitions do not contain class tags. By doing so, the actual on-demand composition is only required for the states in personalized FST. For a multi-threaded service, the pre-composed FST can be shared by all threads, since it does not contain personalized FST states (non-terminals). The personalized part will be shared for all utterances from the same user, which will take full advantage of memory usage. 
	
	Unlike the previous pre-initialization approach that is based on calculating the state statistics~\cite{allauzen2013pre}, our simplified pre-initialization methods do not rely on pre-calculated state frequencies. Instead, we directly expand the graph with breadth-first search or through a data-driven approach where a small numbers of utterances are processed by the decoder offline. We found that both methods are effective, but the data-driven approach outperforms the breadth first search algorithm. Both methods can be combined to achieve the best performance. Through a series of experiments on a speech recognition task for the calling domain, we found that pre-initialization on the public graph speeds up the decoding time by a factor of three. Futhermore, sharing the private graph further reduces decoding time and results in factor of five improvement in efficiency.
	
	\section{Architecture and Algorithm}
	
	The general composition algorithm is well-explained in~\cite{allauzcompose, oonishicompose} and a pre-composition algorithm with a non-class LM is described  in~\cite{allauzen2013pre}. Here we will only present our new algorithm focusing on how to pre-compose the graph while avoiding non-terminal states. In this work, we use the same mathematical notation as \cite{mohriwfst}. 
	
	\subsection{Two-layer cached FST  during decoding}
	
	A WFST can be written as
	\begin{equation}
	T = (\mathcal{A}, \mathcal{B}, Q, I, F, E, \lambda, \rho)
	\end{equation}
	where $\mathcal{A}$, $\mathcal{B}$ are finite label sets for input and output. $Q$ is the finite state set. $I\subseteq Q$ is the initial state set, $F\subseteq Q$ is final state set. $E\subseteq Q\times (\mathcal{A} \cup \{\epsilon\}) \times (\mathcal{B} \cup \{\epsilon\}) \times \mathbb{K} \times Q$ is a set of transitional mapping between states in $Q$ with weighted input/output label pair, where $\mathbb{K}$ is a semiring $(\mathbb{K}, \oplus, \otimes, \overline{0}, \overline{1})$. 
	
	The composition of two weighted FSTs is defined as
	\begin{equation}
	(T_1 \circ T_2)(x, y) = \bigoplus_{z \in\mathcal{B} } T_1(x, z) \otimes T_2(z, y) 
	\end{equation}
	where $\mathcal{B} = \mathcal{B}_1 \cap \mathcal{A}_2$ is the intersection of output label set of $T_1$ and input label set of $T_2$. 
	For $a, b, c\neq \epsilon$, two transitions $(q_1, a, b, w_1, q_1')$ in $T_1$ and $(q2, b, c, w_2, q_2')$, the composed transition will be $((q_1, q_2), a, c, w_1 \bigotimes w_2, (q_1', q_2'))$. 
	
	For two FSTs $T_1$, $T_2$ over semiring $\mathbb{K}$, 
	\begin{equation}
	T_1 = H  C  L
	\end{equation}
	\begin{equation}
	T_2 = Replace(G_c, G_p), p \in \mathcal{C}
	\end{equation}
	is the class language model transducer obtained by replacing the class labels in generic root FST $G_c$ with class FSTs $G_p$ for different classes, where $\mathcal{C}$ denotes the set of all supported classes. 
	
	The calculation for composition is very slow for LM with large vocabulary size. Naive on-the-fly composition is very time-consuming. In \cite{allauzen2013pre}, the authors proposed a pre-initialized composition algorithm, which does a partial composition based on the state frequency. This one-time cost calculation can do some composition in advance. During decoding search, the FST will skip the composition of pre-initialized states. However, extending this algorithm to class LMs is non-trivial in practice. For a class LM, the non-terminal states cannot be composed during pre-initialization since we need a pre-initialization that is applicable to all users, which means we need to apply some restrictions to prevent composition of the personalized part.  
	
	We define $T_P$ as a partial composed FST structure for $T=T_1 \circ T_2$, where $P \subseteq Q$ is the set of pre-composed states. In real time decoding, the on-the-fly composition will be performed on top of the pre-initialized $T_P$, which is similar to previous work \cite{allauzen2013pre}. In a production environment, multiple threads will share the same pre-composed FST $T_P$ structure, while each thread will own a private FST structure.
	\begin{equation}
	T = T_P \cup T_D
	\end{equation}
	where $T_D$ is the dynamic cache built on top of $T_P$. $T_D$ may need to copy some states from $T_P$ if we need to update information for those states in $T_P$. 
	
	\begin{figure}[htb]
		
		\begin{minipage}[b]{1.0\linewidth}
			\centering
			\centerline{\includegraphics[width=5cm]{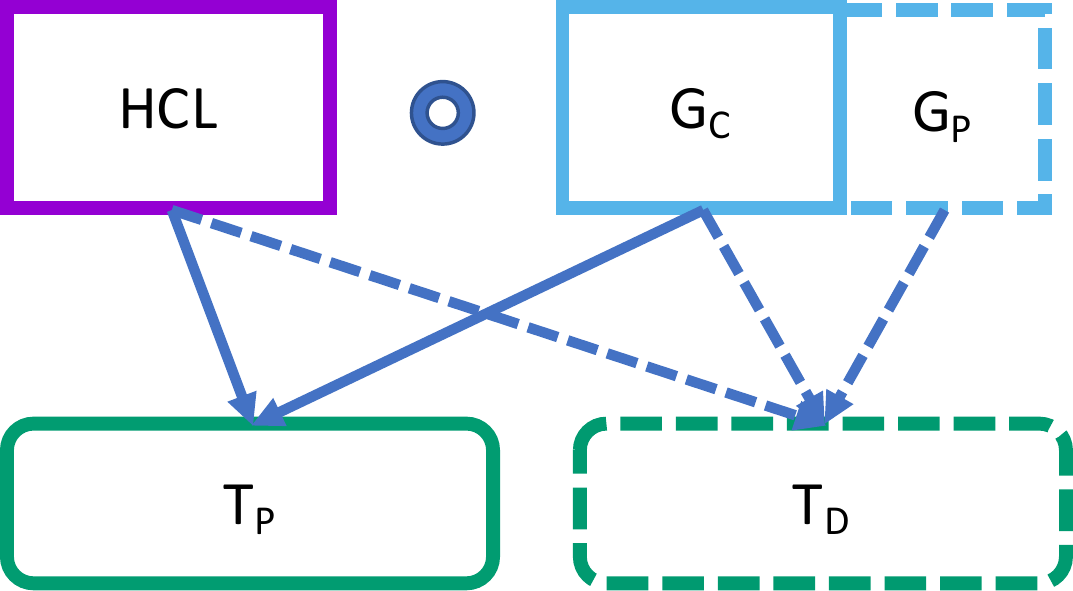}}
		\end{minipage}
		\caption{Architecture of two layer cached FST. $T_P$ is the static public cache built in pre-initialization. $T_D$ is the private cache for dynamic composition in decoding time. The lifetime of $T_D$ varies based on the length of dialog section. }
		\label{fig:draw}
	\end{figure}

	In order to support this mechanism, we use a two-layered cached FST for decoding. The first layer is public cache which represents $T_P$. It is a static cache created by pre-initialization. The second layer is the private cache, which is owned by a particular user and constructed on-the-fly. Figure \ref{fig:draw} shows the architecture of our two-layer FST. The solid box denotes the static graph and the dashed ones show the dynamic graph. Personalized states will appear only in $T_D$.
	
	The static public cache stores the most frequent states, which greatly reduces the run time factor (RTF) of online decoding. Since $T_D$ has a smaller size than a fully dynamic graph, the marginal memory efficiency for multi-threaded service will be better.
	
	Furthermore, the private cache will not be freed after decoding a single utterance. The lifetime of a private cache actually can last for the entire dialog section for a specific user. The private cache keeps updating during the dialog session, making processing the subsequent utterances faster as more states are composed and stored in $T_D$. With this accumulated dynamic cache, a longer dialog can expect a better RTF in theory. In general, the static public cache serves all threads, while the private cache boosts the performance within a dialog session. The private cache will be freed at the end of the dialog. 
	
	\subsection{Pre-composition algorithm for class language models}
	
	Based on the algorithm described in~\cite{allauzen2013pre}, we allow the states $(q_1, q_2)$  such that  $q_2 = (q_c, q_p), q_c \in Q_c, q_p=0 $ to be pre-composed, where $q_c$ and $q_p$ denote states in $G_c$ and $G_p$, respectively. States in $G_c$ with a class label transition will be ignored during pre-composition. 
    \begin{equation}
	P = \cup \{(q_1, (q_c, 0 ))\},  \forall e \in E(q_c): o(e)\notin \mathcal{C}
	\end{equation}
	By applying this restriction, the states in the pre-composed recognition FST $T_P$ will not contain any personalized states, and thus, can be shared by all users and threads.  
	
	%

\begin{figure}[htb]
		
		\begin{minipage}[b]{1.0\linewidth}\label{figure:a}
			\centering
			\centerline{\includegraphics[width=7cm]{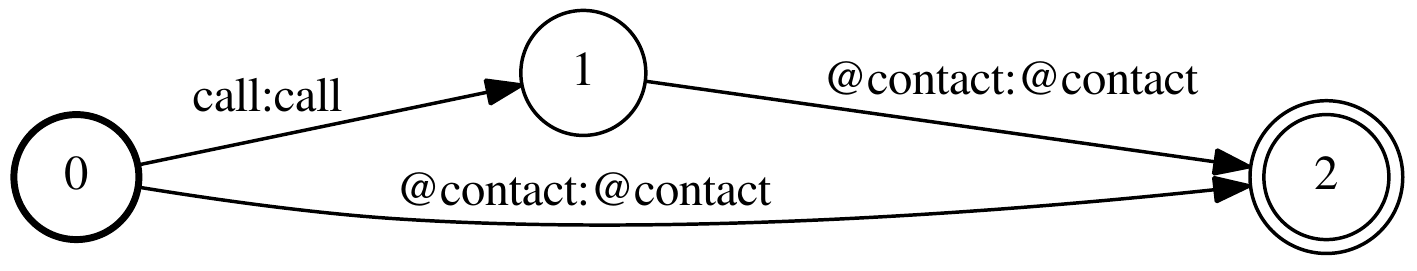}}
			\centerline{(a) }\medskip
		\end{minipage}
		\begin{minipage}[b]{1.0\linewidth}\label{figure:b}
			\centering
			\centerline{\includegraphics[width=7cm]{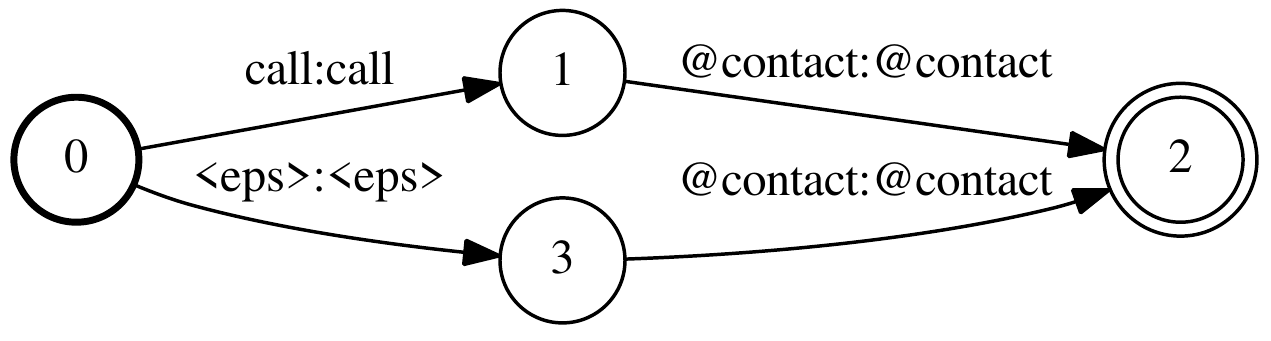}}
			\centerline{(b) }\medskip
		\end{minipage}
		\caption{Class language model FST with contact tags. (a) Conventional LM with @contact. (b) LM with additional $<$eps$>$ between start state 0 and @contact. This guarantees the start state is pre-composed. }
		\label{fig:lm}
	\end{figure}

	Note that care must taken to account for the special case when the initial states could have transitions with a class label. In this case, the entire graph is blocked (Figure~\ref{fig:lm}(a)), so we need to add an extra $\epsilon$ transition before class label in the root FST, which will guarantee all the initial states are composed (Figure~\ref{fig:lm}(b)). In the pre-composition stage, we don't need the actual class FSTs for each class, so $G_p$ is simply a placeholder FST which only contains a placeholder word $\left\langle temp \right\rangle$. This means all the transitions following the placeholder transition may be blocked if there is no other path that skips over the placeholder  transition. In practice, for a large LM graph with a large vocabulary, the connectivity is usually very high, once the initial states are guaranteed to be composed. 
	
	This pre-composition algorithm can be applied with lookahead filter~\cite{novakjlookahead}. We implemented this algorithm using OpenFst framework~\cite{openfst}, which supports such a lookahead filter in both the pre-composition and decoding stages. In our implementation, the decoding FST has a two-layered cache and state table. The state table is necessary since the add-on composition during decoding must be based on the same state map.
	
	\subsection{Pre-composition methods}
	
	In general, we can pre-compose all the states of the decoding FST that are applied to all users, i.e. those unrelated to the personalized language model. However, this full set pre-composition could be very slow and memory consuming. In fact, most of the states are rarely composed during real data traffic, and therefore, performing partial pre-composition is sufficient. Here we propose two simple methods for pre-composition.
	
	\subsubsection{Distance based method}
	
	Naive breath-first-search (BFS) is the most obvious way to perform pre-composition. We iterate over all states within a specific distance from the start state of decoding FST. It generalizes to a full set pre-composition when the search depth is large.
	
	\subsubsection{Data-driven warm-up}
	
	Our goal is to pre-compose the most frequently encountered states. However, if some frequent states are far from the start state, they may not be identified by naive BFS. In this case, it is very time and memory consuming to increase the depth of the BFS. Moreover, if we simply use a offline corpus of utterances to analyze the frequency of all states, some highly frequent states could be blocked by less frequent states. Thus, the easiest way is to do pre-composition using real utterances.  
	
	The decoding FST can be expanded while decoding utterances. We utilize a special decoder in the warm-up stage. This warm-up decoder will apply the same restriction discussed in the previous section. We use an empty contact FST in the warm-up stage to avoid expanding any personalization-related states. This data driven pre-composition will expand most frequent states which are visited during warm-up decoding, especially for some specific patterns.

	\subsection{Out-Of-Vocabulary recognition}
	
	Handling out-of-vocabulary (OOV) words in speech recognition is very important especially for contact name recognition. We replace the normal class (contact) FST with a mono-phone FST by adding monophone words in the lexicon~\cite{aleksic2015improved, dixon2012specialized,McGraw_2016}. By using s monophone FST, we avoid the necessity of adding new words into lexicon on-the-fly, which significantly simplifies the system. We use silence phone "SIL" to represent the word boundary. These monophone words will not be applied with silence phone in lexicon since they are not real words. 
	
\begin{figure}[htb]
		
		\begin{minipage}[b]{1.0\linewidth}
			\centering
			\centerline{\includegraphics[width=8.5cm]{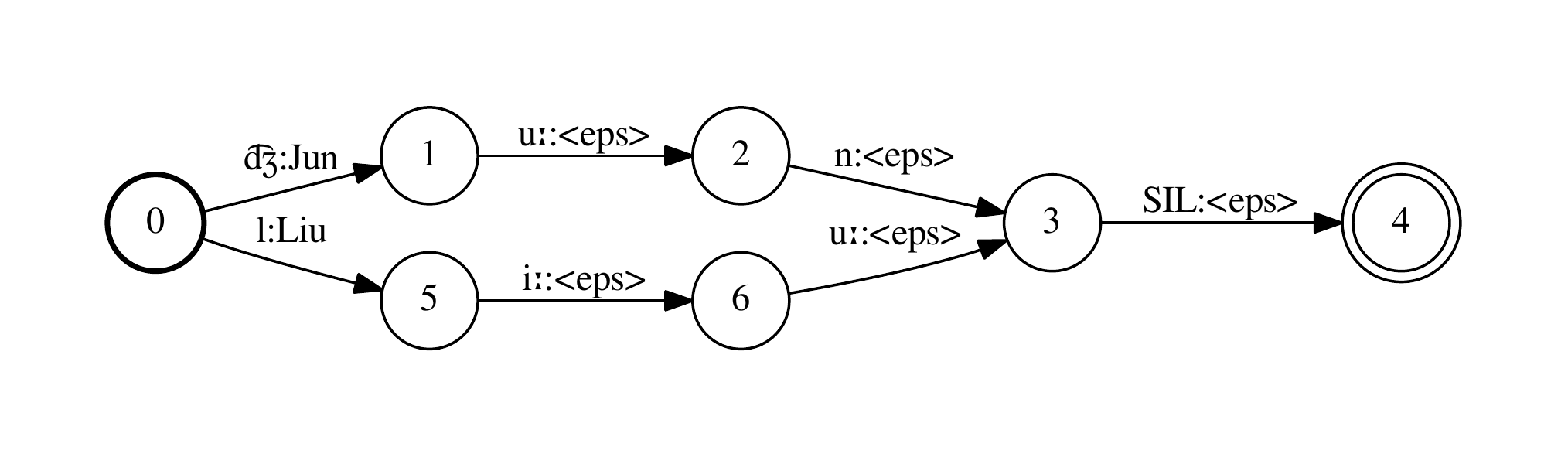}}
		\end{minipage}
		\caption{Monophone contact FST. The monophone will be treated as word in the lexicon without a word boundary, so there is an additional silence phone after each name. }
		\label{fig:plm}
	\end{figure}

	In Figure~\ref{fig:plm}, the contact name is represented as monophone words using IPA phone set. SIL is added after each name in contact FST. Names with the same pronunciation also need to be handled using disambiguation symbols. In practice, because of accent and pronunciation variability, we have found that multiple pronunciations of OOV names are required in the personalized class FST.
	
	\section{Experiments}
	
	We performed a series of experiments on different data sets in order to evaluate the impact on real-time factor (RTF) and word error rate (WER) of the proposed approach. In theory, the pre-composition algorithm will not change the WER, since the search algorithm does not change.
	
	\subsection{Experimental Setup}
	
In these experiments, speech recognition was performed using a hybrid LSTM-HMM framework. The acoustic model is an LSTM that consumes 40-dimensional log filterbank coefficients as the input and generates the posterior probabilities of 8000 tied context-dependent states as the output. The LM is a pruned 4-gram model trained using various semantic patterns that include a class label as well as a general purpose text corpus. The LM contains $@contact$ as an entity word, which will be replaced by the personalized contact FST. After pruning, the LM has 26 million n-grams. 
	
	The personalized class FST (contact FST) only contains monophone words. Determinization and minimization are applied to the contact FST with disambiguation symbols. The disambiguation symbols are removed after graph optimization. The decoding experiments are performed on a server with 110 GB memory and 24 processors.  
	

	\begin{table*}[tbh]
		\centering
		\caption{WER and RTF results for different data set and different pre-composition methods.}
		\label{tab:rtf}
		\begin{tabular}{ cl l l l l  r }
			\toprule
			\multicolumn{1}{c}{\textbf{Dataset}} & 
			\multicolumn{1}{c}{\textbf{Graph Type}} & 
			\multicolumn{1}{c}{\textbf{\# of Pre-composed States}} & 
			\multicolumn{1}{c}{\textbf{RTF(p50)}} & 
			\multicolumn{1}{c}{\textbf{RTF(p95)}} & 
			\multicolumn{1}{c}{\textbf{WER}} \\
			\midrule
			&fully dynamic & 0 & 0.887 & 1.44 &5.48~~~             \\
			calling                        & 5 steps BFS & 1,035,374   & 0.451 &0.793 &5.48~~~       \\
			& data driven warmup & 19,356,186 & 0.286 & 0.484 &5.48~~~              \\
			\midrule
			& fully dynamic & 0  & 0.851  & 1.47 & 7.39~~~               \\
			non-calling                   & 5 steps BFS & 1,035,374 & 0.402 &0.736 & 7.39~~~              \\
			& data driven warmup  & 19,356,186 &0.241 & 0.488 & 7.39~~~              \\
			\bottomrule
		\end{tabular}
		
	\end{table*}

	Experiments are performed on two data sets. The first contains 7,500 utterances from the calling domain from Facebook employees. This includes commands like ``Please call {\it Jun Liu} now". The second consists of approximately 10,000 utterances from other common domains, such as weather, time, and music. Note that we include the contact FST for both calling and non-calling utterances, as we do not assume knowledge of the user's intent a priori. Each user has a contact FST containing 500 contacts on average. We keep up to five pronunciations for each name, generated by a grapheme-to-phoneme model. 
	
	We experiment with both the naive BFS and the proposed data-driven pre-composition methods. For the data-driven approach, we randomly picked 500 utterances from the evaluation data set as warm up utterances. We use an empty contact FST to be replaced into the root LM to avoid personalized states during warm-up decoding. In order to evaluate the benefit of the proposed private cache to store the personalized language model, we group multiple utterances from a user into virtual dialog sessions of one, two, or five turns. 
	
	\subsection{Results}
	
	\begin{table}[tbh]
		\centering
		\caption{RTF results for decoding in session. Decoder will hold the private cache for entire dialog session.}
		\label{tab:rtf_private}
		\begin{tabular}{ c  l c l  l  r }
			\toprule
			\multicolumn{1}{c}{\textbf{Dataset}} & 
			\multicolumn{1}{c}{\textbf{Session Length}} & 
			\multicolumn{1}{c}{\textbf{RTF(p50)}} & 
			\multicolumn{1}{c}{\textbf{RTF(p95)}} \\
			\midrule
			& 1 & 0.286 & 0.484~~~             \\
			calling &   2 & 0.220 & 0.390~~~       \\
			& 5 & 0.182 & 0.346~~~          \\   
			\midrule
			& 1 & 0.241 & 0.488~~~             \\
			non-calling &   2 & 0.203 & 0.408~~~       \\
			& 5 & 0.173 & 0.372~~~          \\   
			\bottomrule
		\end{tabular}
		
	\end{table}
	
	\begin{figure}[!tbh]
		
		\begin{minipage}[b]{1.0\linewidth}
			\centering
			\centerline{\includegraphics[width=7.8cm]{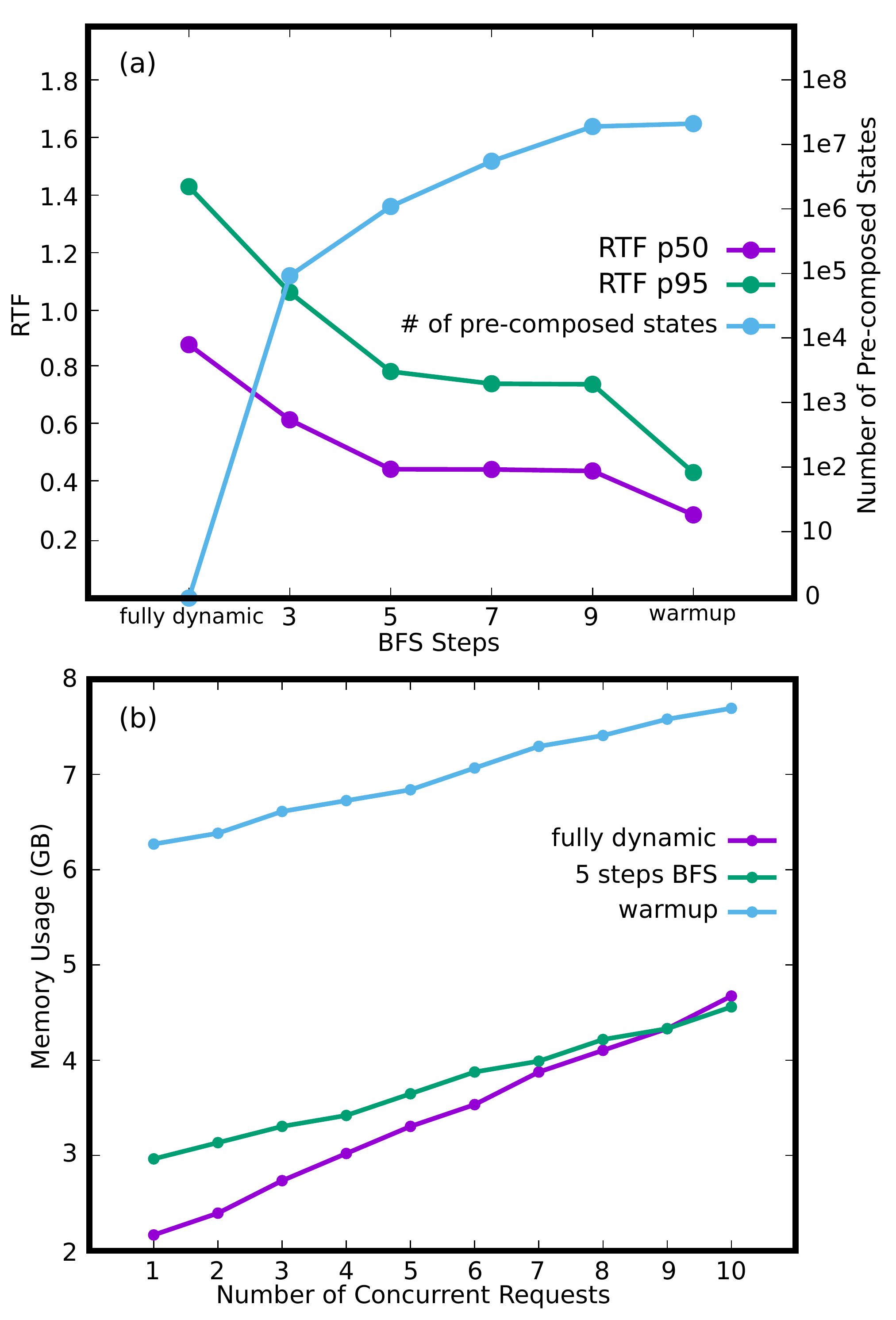}}
		\end{minipage}
		\caption{RTF and memory usage comparison. Upper: RTF between fully dynamic graph, different steps of BFS and data driven pre-composition. Lower: Memory usage for different graphs. A pre-composed graph has a better marginal memory cost than a fully dynamic graph.}
		\label{fig:result}
	\end{figure}
	
	Table~\ref{tab:rtf} shows the WER and RTF for two corpora with different pre-composition methods with ten concurrent speech recognition client requests. The private cache is freed after decoding each utterance. RTF is calculated by $t_{decode}/t_{wav}$, where $t_{decode}$ is the decoding time and $t_{wav}$ is the audio duration. We use 50th and 95th percentile values for the RTF comparison. As expected, the WER remains unchanged for the same data set. With pre-composition, the RTF for both calling and non-calling is reduced by a factor of three. 
	
	Table~\ref{tab:rtf_private} shows the additional RTF improvement that can be obtained during multi-turn dialogs from the proposed private cache. When the dialog session is only a single turn, the RTF remains unchanged. However, for multi-turn sessions, additional RTF reductions are obtained for both the calling and non-calling corpora. The decoding time is reduced by a factor of five compared to a fully dynamic graph for dialog sessions of five turns. 
	
	Figure~\ref{fig:result} shows the RTF and memory usage for teh different pre-composition approaches. The upper graph shows the RTF for different steps of naive BFS using the calling data set. The figure shows that additional BFS steps improves RTF for both 50 and 95 percentiles. However, no improvement is observed beyond five steps, because the most frequent states close to the start state have already been pre-composed. The additional BFS steps only result in more memory usage. With the data-driven warmup, the RTF shows additional improvement. Furthermore, the difference in the p50 and p95 RTF values becomes much smaller than in the BFS approach.  
	
	The lower graph of Figure~\ref{fig:result} shows the memory usage as a function of the number of concurrent requests. Though the pre-composed graph may use more memory when we have only a small number of threads, the marginal memory cost for additional requests for a fully dynamic graph is roughly 1.5 times larger than for the pre-composed graph. The data-driven method has the best marginal memory efficiency for a large number of concurrent requests.
	
	\section{Conclusions}
	
	In this work, we propose new methods for improving the efficiency of dynamic WFST decoding with personalized language models. Experimental results show that using a  pre-composed graph can reduce the RTF by a factor of three compared with a fully dynamic graph. Moreover, in multi-utterance dialog sessions, the RTF can be reduced by a factor of 5 using the proposed private cache without harming WER. Though a fully dynamic graph uses less memory for the graph, the pre-composed graph has a better marginal memory cost, which is more memory efficient in large-scale production services that need to support a large number of concurrent requests.  
	
	Our results also show that increasing the steps of naive BFS will not help the RTF, since it may compose infrequently encountered states, resulting in unnecessary memory usage. Using the proposed data-driven warm-up performs better in both marginal memory efficiency and RTF than naive BFS. Both pre-composition methods can also be combined.

\section{Acknoledgements}

We would like to thank Mike Seltzer, Christian Fuegen, Julian Chan, and Dan Povey for useful discussions about the work.
	
	\bibliographystyle{IEEEtran}
	
	\bibliography{mybib}

\begin{thebibliography}{10}
\providecommand{\url}[1]{#1}
\csname url@samestyle\endcsname
\providecommand{\newblock}{\relax}
\providecommand{\bibinfo}[2]{#2}
\providecommand{\BIBentrySTDinterwordspacing}{\spaceskip=0pt\relax}
\providecommand{\BIBentryALTinterwordstretchfactor}{4}
\providecommand{\BIBentryALTinterwordspacing}{\spaceskip=\fontdimen2\font plus
\BIBentryALTinterwordstretchfactor\fontdimen3\font minus
  \fontdimen4\font\relax}
\providecommand{\BIBforeignlanguage}[2]{{%
\expandafter\ifx\csname l@#1\endcsname\relax
\typeout{** WARNING: IEEEtran.bst: No hyphenation pattern has been}%
\typeout{** loaded for the language `#1'. Using the pattern for}%
\typeout{** the default language instead.}%
\else
\language=\csname l@#1\endcsname
\fi
#2}}
\providecommand{\BIBdecl}{\relax}
\BIBdecl

\bibitem{mohriwfst}
M.~Mohri, F.~Pereira, and M.~Riley, ``Speech recognition with weighted
  finite-state transducers,'' \emph{Handbook of Speech Pro-cessing, Springer},
  pp. 559--582, 2008.

\bibitem{mowfst}
------, ``Weighted finite-state transducersin speech recognition,''
  \emph{Computer Speech \& Language}, vol.~20, no.~1, pp. 69--88, 2002.

\bibitem{aleksic2015improved}
P.~Aleksic, C.~Allauzen, D.~Elson, A.~Kracun, D.~M. Casado, and P.~J. Moreno,
  ``Improved recognition of contact names in voice commands,'' in
  \emph{ICASSP}, 2015, pp. 5172--5175.

\bibitem{allauzen2013pre}
C.~Allauzen and M.~Riley, ``Pre-initialized composition for large-vocabulary
  speech recognition,'' in \emph{INTERSPEECH}, 2013, pp. 666--670.

\bibitem{openfst}
C.~Allauzen, M.~Riley, J.~Schalkwyk, W.~Skut, and M.~Mohri, ``{OpenFst: A
  general and efficient weighted finite-state trans-ducer library},'' in
  \emph{CIAA}, 2007, pp. 11--23.

\bibitem{allauzcompose}
C.Allauzen, M.Riley, and J.Schalkwyk, ``A generalized composition algorithm for
  weighted finite-state transducers,'' in \emph{INTERSPEECH}, 2009, pp.
  1203--1206.

\bibitem{oonishicompose}
T.~Oonishi, P.~Dixon, K.~Iwano, and S.~Furui, ``{Implementation and evaluation
  of fast on-the-fly WFST composition algorithms},'' in \emph{INTERSPEECH},
  2008, pp. 2110--2113.

\bibitem{novakjlookahead}
J.~R. Novak, N.~Minematsu, and K.~Hirose, ``{Dynamic grammars with lookahead
  composition for WFST-based speech recognition},'' in \emph{ICASSP}, 2012.

\bibitem{dixon2012specialized}
P.~R. Dixon, C.~Hori, and H.~Kashioka, ``{A specialized WFST approach for class
  models and dynamic vocabulary},'' in \emph{INTERSPEECH}, 2012.

\bibitem{McGraw_2016}
I.~McGraw, R.~Prabhavalkar, R.~Alvarez, M.~G. Arenas, K.~Rao, D.~Rybach,
  O.~Alsharif, H.~Sak, A.~Gruenstein, F.~Beaufays, and C.~Parada,
  ``{Personalized speech recognition on mobile devices},'' in \emph{ICASSP},
  2016.

\end{thebibliography}
	
\end{document}